# Bridging the Gap between Language Models and Cross-Lingual Sequence Labeling

**Nuo Chen[1]**[*], **Linjun Shou[2]**, **Ming Gong[2]**, **Jian Pei[3]**, **Daxin Jiang[2]**[†]

[1]ADSPLAB, School of ECE, Peking University, China
[2]STCA NLP Group, Microsoft, Beijing
[3]School of Computing Science, Simon Fraser University

nuochen@pku.edu.cn, {lisho,migon,djiang}@microsoft.com, jpei@cs.sfu.ca

## Abstract

Large-scale cross-lingual pre-trained language models (xPLMs) have shown effectiveness in cross-lingual sequence labeling tasks (xSL), such as cross-lingual machine reading comprehension (xMRC) by transferring knowledge from a high-resource language to low-resource languages. Despite the great success, we draw an empirical observation that there is a training objective gap between pre-training and fine-tuning stages: e.g., mask language modeling objective requires *local* understanding of the masked token and the span-extraction objective requires *global* understanding and reasoning of the input passage/paragraph and question, leading to the discrepancy between pre-training and xMRC. In this paper, we first design a pre-training task tailored for xSL named Cross-lingual Language Informative Span Masking (CLISM) to eliminate the objective gap in a self-supervised manner. Second, we present ContrAstive-Consistency Regularization (CACR), which utilizes contrastive learning to encourage the consistency between representations of input parallel sequences via unsupervised cross-lingual instance-wise training signals during pre-training. By these means, our methods not only bridge the gap between pretrain-finetune, but also enhance PLMs to better capture the alignment between different languages. Extensive experiments prove that our method achieves clearly superior results on multiple xSL benchmarks with limited pre-training data. Our methods also surpass the previous state-of-the-art methods by a large margin in few-shot data settings, where only a few hundred training examples are available.

## 1 Introduction

Sequence labeling (SL) tasks are common conditions and have considerable impact in natural language processing communities, such as named entity recognition (NER) (Lample et al., 2016), and machine reading comprehension (MRC) (Hermann et al., 2015; He et al., 2018; Asai et al., 2018; Yuan et al., 2020; You et al., 2020b; You et al.), to name a few. For instance, MRC models are required to predict whether each token in the input passage is the start or end position of the ground-truth answer span given a pre-defined question. Many promising results have been achieved using deep-learning based methods (Seo et al., 2017; Liang et al., 2021b) on popular benchmarks (Rajpurkar et al., 2016, 2018; Reddy et al., 2019; You et al., 2020a). Recently, there is a growing body of literature that recognizes the importance of cross-lingual sequence labeling tasks (xSL for short) (Huang et al., 2019; Lewis et al., 2020; Artetxe et al., 2019a), where translation systems are utilized to translate high-resource languages into low-resource languages so as to enrich the training data. However, the performance of xSL models is severely affected by translation quality (Yuan et al., 2020; Liang et al., 2021a).

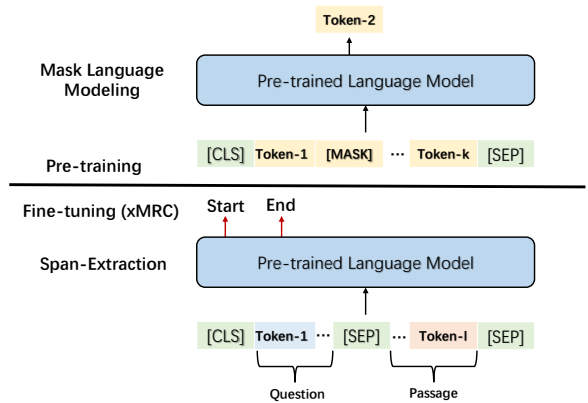

Figure 1: Mask language modeling objective in the pre-training stage and span-extraction objective of xMRC in the fine-tuning stage.

Recently, thanks to multilingual pre-trained language models (xPLMs) (Huang et al., 2019; Liang et al., 2020; Conneau et al., 2019; Chi et al., 2021a) pre-trained on large multilingual corpus (billion-level), the pre-training + fine-tuning paradigm starts dominating various cross-lingual understanding tasks including entity recognition (Liang et al., 2021a,b) and question answering (Asai et al., 2018; Chen et al., 2021b). Even if in zero-shot or few-shot settings, these effective xPLMs still achieve remarkable improvements during fine-tuning via transfer learning. Despite the promising results in such settings, there is an objective gap between pre-training and fine-tuning for xSL tasks. Take the xMRC task as an example. During pre-training, PLMs are always trained with mask language modeling (MLM) (Devlin et al., 2018; Lan et al., 2020), Next Sentence Prediction (NSP) (Pires et al., 2019; Conneau et al., 2019) or translation replaced token detection (TRTD) (Chi et al., 2021c) tasks. Subsequently, these state-of-the-art models are typically trained for span extraction using distantly supervised multilingual task-related examples during fine-tuning. This is somewhat expected, since *local* context understanding around the mask token is mainly needed to perform pre-training tasks like MLM or TRTD, while span-extraction MRC requires the model to understand the interaction between question and *global* context of passage, as shown in Figure 1. Suffering from different objectives, the performance of the model inevitably deteriorates (Jiao et al., 2021; Ram et al., 2021).

---
[*]Work done when interned at Microsoft STCA.
[†]Corresponding Author

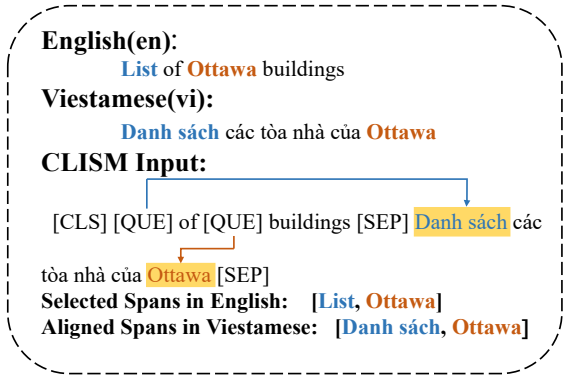

Figure 2: The example of our proposed CLISM. Here we present the input sequence in English and Viestamese. Selected spans in English are from off-the-shelf NER tools like Spacy and the corresponding aligned spans in Viestamese obtained by GIZA++ toolkit (Pei et al., 2020).

Motivated by these factors, we first design a cross-lingual language-informative span masking (CLISM) task for bridging the gap between pre-training and fine-tuning in the context of xSL, which can be seen as a question answering task. A primary concern of such a self-supervised task is how to create corresponding multilingual <question, answer> training pairs. Concretely, we group the parallel corpus into several sub-groups. Each sub-group is called a language-informative group and contains the same sentence in two different languages, a source language and a target language. Considering aligned spans in the cross-lingual parallel sentences can be seen as a signal for coreference and lead to information short-cut (Jiao et al., 2021). Subsequently, we mask n-grams (e.g., named entities) spans which occur in a given source language sentence, and then we take the *masked* sentence and the unmasked counterpart in target language as input. Thus, we form question answering by masking all selected spans with a special token [QUE], and requiring the model to find the correct start and end positions for each of such tokens based on understanding global context of the sentence in source and target languages, as shown in Figure 2. Correspondingly, each [QUE] is regarded as a *question*, and its corresponding span in the target language sentence that is unmasked acts as the *answer*. With this strategy, the model needs to predict whether each token is the answer start or end index, which is essentially consistent with xSL tasks (Ram et al., 2021). Hence the training objectives are connected, and thus, cross-lingual PLMs can be more adequate for xSL.

Considering to better capture the alignment between the same content in different languages and avoid learning representations that are covariant with the noisy parallel input (e.g., [QUE] and [MASK]), we also design a strategy called ContrAstive-Consistency Regularization (CACR) during pre-training. Our core idea behind CACR is to push representations of the same sequence in various languages to be similar, while pushing it away from other examples via contrastive learning. The overview of our architecture is presented in Figure 3.

The resulting language model exhibits surprisingly performances in multiple xSL benchmarks, such as xMRC and xNER. Of note, compared with other new state-of-the-art xPLMs such as Info-XLM and XLM-Align which utilize billion-level pre-training data, our model pre-trained with limited data e.g., 1 million achieve better performances across various datasets. As an additional benefit of our methods, our model also achieves superior performances given a few training instances. For example, ours achieve 35.9%/50.8% EM/F1 score with only 512 examples on MLQA dataset, outperforming all baselines by a very wide margin (vs. 15.5%/26.8%).

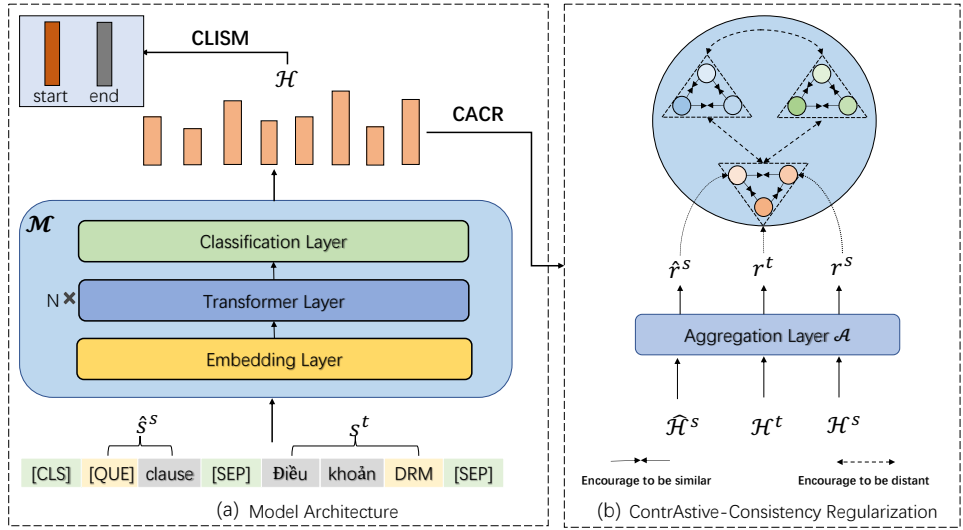

Figure 3: Model Architecture: In this example, we choose English (en) as our source language and Viestamese (vi) as our target language, separately. In the input sequence, the original sentence in English is "DRM clause", "DRM" is the selected span. We mask "DRM" with the special token [QUE]. Then the designed CLISM is required the model to predict the start and end positions of the correct answer for [QUE] ("DRM" in target language).

In general, we make the following technical contributions in this paper:

- We present a new cross-lingual pre-training paradigm named CLISM to bridge the training objective gap between language models and cross-lingual sequence labeling tasks.

- We design a novel pre-training strategy termed CACR, which encourages xPLMs to better capture alignment between cross-lingual representations.

- The resulting models achieve state-of-the-art performances with limited parallel pre-training data on multiple xSL benchmarks. Moreover, we demonstrate that our method also shows significant improvements in few shot data settings during fine-tuning.

## 2 Related Work

**Pre-trained Language Models** During the past few years, PLMs (Devlin et al., 2018; Liu et al., 2019; Lan et al., 2020; Sun et al., 2020) dominate natural language understanding and generating areas due to their promising results in various downstream tasks. Taking a step forward a more real world, there is a growing body of communities extent PLMs to multi-lingual language (xPLMs), and some representative works have been done such as XLM-R (Conneau et al., 2019) (short for XLM-Roberta), info-XLM (Chi et al., 2021a) and XLM-Align (Chi et al., 2021b). Almost these efforts trained with *token-level* pre-training tasks on large multi-lingual corpus. Directly utilizing xPLMs lead high performances in the fine-tuning stage, but when transferred to downstream tasks can result the objective gap between pretrain-finetune, making xPLMs sub-optimized in downstream tasks especially for xSL (Jiao et al., 2021; Ram et al., 2021). Our proposed CLISM and CACR not only alleviate the pretrain-finetune discrepancy, but also strong the ability of xPLMs to capture alignment between cross-lingual representations in sentence-level.

**Sequence Labeling** Sequence Labeling tasks are of great significance in web mining, such as named entity recognition (Shen et al., 2021; Cui et al., 2021), relation identification (Akhmouch et al., 2021), event extraction (Lu et al., 2021; Lyu et al., 2021), and even machine reading comprehension (Wu et al., 2021; Chen et al., 2021a; You et al., 2021a,b; Chen et al., 2021c). However, when pushing the boundary of SL to low-resource languages, xSL can be very challenging, due to limited training data. To tackle this challenge, various efforts have been proposed (Huang et al., 2019; Liang et al., 2020; Conneau et al., 2019; Liang et al., 2021a). Cui et al. (2019) and Singh et al. (2019) utilized machine translation to get parallel data as the data argumentation method. But the performance of these methods are restricted to the quality of translation. Hence, Yuan et al. (2020) proposed to generate low-resource language weakly labeled data from Wikipedia articles. Similarly, we also collect pre-training parallel data for CLISM from web sources. We tackle the objective gap between language models and xSL tasks, an challenge overlooked by the previous methods.

**Recent Works** Currently, several studies have explored to bridge the gap between pre-training and fine-tuning, including two main categories: (1) *knowledge-driven*, which introduces external commonsense knowledge into the pre-training for improving model performance in down-stream tasks (Varkel and Globerson, 2020; Sun et al., 2020), (2) The latter is *task-oriented*, which includes several well designed pre-training or fine-tuning tasks (Jiao et al., 2021; Deng et al., 2021). However, none of these works target cross-lingual areas, especially for xSL tasks. Recently, concurrent with our research, Ram et al. (2021) studied aligning the pre-training and fine-tuning stages in the context of question answering via introducing a novel pre-training task. Nevertheless, Ram et al. (2021) focused on few-shot question answering, and didn't explore the relation between different languages.

## 3 Methodology

In this section, we first illustrate our two pre-training strategies: CLISM and CACR in detail. Then we introduce how to apply our methods in down-stream tasks to obtain high performances. Due to the excellent versatility of our proposed methods, which can build on top of any xPLMs. Hence, we leverage $\mathcal{M}$ to represent a series of pre-trained language models. Our goal aims at aiding $\mathcal{M}$ to eliminate the influence of objective gap between pretrain-finetune, and enhancing the consistency between parallel data representations.

### 3.1 Cross-lingual Language Informative Span Masking

**Definition** Heading for bridging the objective gap between pre-training and fine-tuning in xSL, we formulate a new pre-training task named Cross-lingual Language Informative Span Masking (CLISM) from the unlabeled data. Given an input sentence $\mathbf{s}^s$ in source language and its counterpart in the target language $\mathbf{s}^t$. We first utilize name entity recognition tools like Spacy to select entities[1] in source language. Considering the translation of the same phrase may be different when pushing it into translation systems separately or concat it with a whole sentence. Therefore, we leverage the off-the-shelf alignment tool like GIZA++ (Pei et al., 2020) to align the corresponding words of the selected entity in the target language. After extracting entity spans and alignment[2], we replace these selected spans with a special token [QUE] in source language sentence $\mathbf{s}^s$, and the corresponding alignment spans in $\mathbf{s}^t$ act as the "ground-truth answer" span for each masked "question" ([QUE]). For instance, as the example presented in the Figure 3 (a), the span "DRM" is the selected span, we replace it in source language with [QUE] and utilize the aligned span "DRM" in target language as the correct answer.

**Pre-training** After getting identical language-informative spans $\mathcal{S}$ in $\mathbf{s}^s$, where each of $\mathcal{S}$ is replaced with a single [QUE] token Q. Hence, we can get the final input sequence $\mathbf{X}$ via concating the masked source language sentence $\hat{\mathbf{s}}^s$ and unmasked target language sentence $\mathbf{s}^t$ with two special tokens [SEP] and [CLS], as shown in Figure 3 (a). Concretely, the goal of CLISM can be seen as to predict the correct answer (corresponding aligned span in $\mathbf{s}^t$) for a given Q $\in \mathcal{S}$ based on understanding the meaning of input $\mathbf{X}$. Subsequently, our method converts unlabeled text into a set of specific questions that need to be answered simultaneously. Thus, in order to find the correct answer span of each Q, we pass $\mathbf{X}$ through $\mathcal{M}$, producing contextualized representations for each token, and then use the representation of Q to compute dynamic start and end vectors $\mathbf{s}_q, \mathbf{e}_q$:

$$\mathbf{s}_q = \mathbf{W}_s \mathbf{x}_q \qquad \mathbf{e}_q = \mathbf{W}_e \mathbf{x}_q \qquad (1)$$

where $\mathbf{W}_s$ and $\mathbf{W}_e$ are learnable parameters and $\mathbf{x}_q$ denotes the representation of Q. Subsequently. we follow the standard span-extraction setting (Huang et al., 2019) to obtain the start and end position probability distributions via computing the inner product of learned vectors $\mathbf{s}_q, \mathbf{e}_q$ with each token representation in $\mathbf{X}$:

---
[1] If the extracted entities of one sentence is none, we will remove it.
[2] We further conduct informative span selection strategy, reducing the bias of the extracted entities. Details can be seen in the Appendix B.

$$\mathcal{P}(s=k|\mathbf{X},\mathtt{Q}) = \frac{\exp(\mathbf{x}_k^T \mathbf{s}_q)}{\sum_j \exp(\mathbf{x}_j^T \mathbf{s}_q)} \quad (2)$$

$$\mathcal{P}(e=k|\mathbf{X},\mathtt{Q}) = \frac{\exp(\mathbf{x}_k^T \mathbf{e}_q)}{\sum_j \exp(\mathbf{x}_j^T \mathbf{e}_q)} \quad (3)$$

Finally, we optimize the model via standard cross-entropy loss:

$$\mathcal{L}_{clism} = -\log \mathcal{P}(s=a_i^s|\mathbf{X}) - \log \mathcal{P}(e=a_i^e|\mathbf{X}) \quad (4)$$

where $a_i$ denotes the aligned span of $i$-th Q in $\mathcal{S}$. Thus, the capability of evidence extraction across different languages is enhanced during pre-training, and will be smoothly transferred to xSL during fine-tuning.

### 3.2 ContrAstive-Consistency Regularization

We design CACR from two points of view: (1) Misalignment between parallel sequences representations still hinders xPLMs advancing in downstream tasks, (2) prior works tend to learn corrupted representations (e.g., [MASK] token) which are covariant with the noise during pre-training (Luo et al., 2021). In contrast, our CACR aims at not only aiding $\mathcal{M}$ in learning noise invariant representations during pre-training, but also promoting $\mathcal{M}$ to better capture alignment between representations of parallel input sequences. Therefore, the proposed CACR could support CLISM to better optimize $\mathcal{M}$ during pre-training.

**Definition** Aforementioned, the input sequence $\mathbf{X} = \{\,[\mathtt{CLS}]\,\hat{\mathbf{s}}^s\,[\mathtt{SEP}]\,\mathbf{s}^t\,[\mathtt{SEP}]\,\}$. We then take $\mathbf{X}$ as input to get the contextual representations of sentences:

$$\mathcal{H} = [\mathcal{H}_{cls}, \hat{\mathcal{H}}^s, \mathcal{H}^t, \mathcal{H}_{sep}] = \mathcal{M}(\mathbf{X}) \quad (5)$$

where $\mathcal{H} \in \mathbf{R}^{l \times d}$, l and d are the max input sequence length and hidden size, separately. $\hat{\mathcal{H}}^s \in \mathbf{R}^{n \times d}, \mathcal{H}^t \in \mathbf{R}^{m \times d}$ denote the representations of $\hat{\mathbf{s}}^s$ and $\mathbf{s}^t$. Also m and n represent the sequence length of $\hat{\mathbf{s}}^s$ and $\mathbf{s}^t$. Note that we can not directly utilize $\hat{\mathcal{H}}^s$ and $\mathcal{H}^t$ as the input of our CACR because of the mismatch between m and n. Thereafter, we apply an extra aggregation layer (e.g., mean-pooling) $\mathcal{A}$ to obtain final global semantics of $\hat{\mathbf{s}}^s$ and $\mathbf{s}^t$:

$$\hat{\mathbf{r}^s} = \mathcal{A}(\hat{\mathcal{H}}^s) \qquad \mathbf{r}^t = \mathcal{A}(\mathcal{H}^t) \quad (6)$$

where $\hat{\mathbf{r}^s}$ and $\mathbf{r}^t$ belong to $\mathbf{R}^d$. Intuitively, the most simple implement of our CACR is to regard $\hat{\mathbf{r}^s}$ and $\mathbf{r}^t$ as positives and others in mini-batch are negatives to optimize the $\mathcal{M}$ via a standard contrastive objective (Hjelm et al., 2019):

$$\mathcal{L}(\mathbf{r}^t, \hat{\mathbf{r}^s}) = -\log \frac{\exp(\Psi(\mathbf{r}^t, \hat{\mathbf{r}^s})/\tau)}{\sum_{j=1}^{B} \exp(\Psi(\mathbf{r}^t, \mathbf{r}_j)/\tau)} \quad (7)$$

where $B$ and $\tau$ are mini-batch and temperature, $\Psi(,)$ denotes the cosine similarity function. Nevertheless, considering $\hat{\mathbf{s}}^s$ is the corrupted input, which contains the [QUE] token. Inspired by (Luo et al., 2021), optimizing our $\mathcal{M}$ only with Eq.7 may lead the model to learn representations that are covariant with the noise text, causing incorrectly represented alignment between source language and target language. Therefore, we further introduce the original unmasked sentence in source language $\mathbf{s}^s$ as input. Similarly, the pre-trained model $\mathcal{M}$ encodes $\mathbf{s}^s$ to obtain the hidden representation $\mathcal{H}^s$. Also the designed aggregation layer $\mathcal{A}$ is responsible for getting the final semantic representation $\mathbf{r}^s$. Thereafter, the positive threefold is collected: $\{\mathbf{r}^s, \mathbf{r}^t, \hat{\mathbf{r}^s}\}$. Each representation in this set is regarded as positive, that is, we expect each two of them should be similar as possible in the latent space. By these means, $\mathcal{M}$ is encouraged to learn noise-invariant representations and better capture cross-lingual representations alignment, alleviating pretrain-finetune discrepancy to some extent, as shown in Figure 3 (b). The final loss of our CACR can be formulated as:

$$\mathcal{L}_{cacr} = \mathcal{L}(\mathbf{r}^t, \hat{\mathbf{r}^s}) + \mathcal{L}(\mathbf{r}^s, \hat{\mathbf{r}^s}) + \mathcal{L}(\mathbf{r}^t, \mathbf{r}^s) \quad (8)$$

### 3.3 Pre-training Strategy

To be consistent with previous methods (Liu et al., 2019; Chi et al., 2021a), we also pre-train $\mathcal{M}$ with the mask language modeling (MLM) training objective. In detail, we train the model in multi-task setting during pre-training. The total objective of our model can be defined as:

$$\mathcal{L} = \mathcal{L}_{clism} + \mathcal{L}_{cacr} + \mathcal{L}_{mlm} \quad (9)$$

### 3.4 Fine-tuning

Inspired by (Ram et al., 2021), to be consistent with our proposed pre-training strategies, we also leverage the special token used in pre-training stage for downstream tasks. In other words, we append the [QUE] token to the input sequence during fine-tuning. Extensive results show this strategy achieve better performances in xSL tasks, especially in few-shot data setting. For example, we can get the following input sequence in xMRC task with adding [QUE] after the input question. :

$$\bar{\mathbf{X}} = \{\texttt{[CLS]}\mathbf{Q}\texttt{[QUE][SEP]}\mathbf{P}\texttt{[SEP]}\} \quad (10)$$

where $\mathbf{Q}$ and $\mathbf{P}$ denote the input question and passage. Due to [QUE] token acts an question during our pre-training. Hence after pre-training, we expect it captures enough information about the question, and thus, we utilize the representation of [QUE] to select the ground-truth answer span in xMRC task. The following process is the same as Section 3.1.

| Data | <en,ar> | <en,hi> | <en,vi> | Total |
|---|---|---|---|---|
| | 0.4M | 0.3M | 0.3M | 1M |

Table 1: Sentence number we used in pre-training.

## 4 Experiments

In this section, we first present the pre-training data collection. Then we illustrate experiments on xSL tasks, including specific implementation and detailed results.

### 4.1 Pre-training Data

Aforementioned, our proposed CLISM and CACR require parallel data. Therefore we utilize MT dataset (Conneau and Lample, 2019) to construct our input sequences. In detail, we only utilize four languages including English(en), Arabic(ar), Vietamese(vi) and Hindi(hi) from MT dataset to train our model, proving our method also achieves great improvements in other languages which we don't obtain any data. Concretely, the inputs of our model consist of two languages: source language and target language. Considering the promising performance of off-the-shelf NER tools (e.g., Spacy) in English, thus we choose English as our source language and another three languages are regarded as the target language in turn. Thereafter, the number of our pre-training data is reported at Table 1, the total amount data is 1 million.

### 4.2 Training Details

**Model Structure** We initialize the parameters of our model from XLM-R and Info-XLM base version published in Hugging Face Transformers[3], showing the generalization of our methods. Hence, the resulting model $\mathcal{M}$ contains 12 transformer layers and hidden state is set to 768.

[3] https://github.com/huggingface/transformers

**Pre-training Details** In our implementation of MLM, we follow the setting in (Devlin et al., 2018), which randomly mask 15% tokens of the input sequence[4]. We optimize our model with the Adam optimizer, using batch size of 64 for a total of 15K steps in pre-training. And learning rate is set to 1e-5 with 1.5K warmup steps. The total max input sequence length is set to 256, that is, the max length of each sentence is set to 128. Notice that $\tau$ in Eq.7 is set to 20. We pre-train our model using $8\times$V100-32G GPUs for 4-5 hours. Fine-tuning details can be seen in the Appendix C.

### 4.3 Evaluation

We conduct experiments over two xSL tasks: xMRC and xNER including four datasets: MLQA (Lewis et al., 2020), XQUAD (Artetxe et al., 2019b), CoNLL (Sang, 2002) and WikiAnn (Pan et al., 2017). Detailed introduction about these datasets presented in the Appendix D.

For all datasets, we fine-tune our model in *zero-shot* setting, which denotes we only utilize the English training set to optimize the model, and test the resulting model on other target languages.

For xMRC task, we use two evaluation metrics, Exact Match (EM) and span-level F1 score, which are popularly used for accuracy evaluation of MRC models. Span-level F1 measures the span overlap between ground-truth answer and model predictions. Exact match (EM) score is 1 if the prediction is exactly the same as the ground truth, otherwise 0. As for xNER task, we use entity-level F1 score to evaluate our model, which requires the boundary and type between the prediction and the ground-truth entity should be matched precisely.

### 4.4 Experiment Results

**Baselines** We compare our model with the following pre-trained baselines: (1) M-BERT (Devlin et al., 2018) pre-trained with MLM task on Wikipedia data in 102 languages; (2) XLM (Conneau and Lample, 2019), another multilingual PLM pre-trained with MLM and TLM tasks in 100 languages; (3) XLM-R (Conneau et al., 2019), a strong effective xPLM pre-trained with MLM in more large corpus; (4) Info-XLM (Chi et al., 2021a), a recent xPLM which continues training XLM-R with MLM, TLM and XLCO in 94 languages; (5) XLM-Align (Chi et al., 2021b), a more recent but effective xPLM, also built on top of XLM-R with MLM and DWA in 94 languages. Besides, we also utilize strong baselines which are specifically designed for xSL tasks: (1) CalibreNet (Liang et al., 2021b) designed for enhancing the boundary detection of xPLMs in xSL tasks; (2) AA-CL (Chen et al., 2021b), a more recent approach to construct hard-negatives via contrastive learning for xMRC. Note that, we reimplement the base version of these baselines in our local environment with 3 times and report the average results of baselines.

[4] The pre-defined answers in CLISM will not be masked to avoid missing labels.

| Models | #data | #languages | MLQA | XQUAD | CoNLL | WikiAnn |
| --- | --- | --- | --- | --- | --- | --- |
| M-BERT | - | 102 | 57.80/42.40 | 69.63/53.72 | 78.2 | 69.1 |
| XLM | - | 100 | 61.70/44.20 | 70.93/53.18 | 79 | 69.5 |
| CalibreNet | - | - | 65.40/48.84 | - | 80.92 | - |
| Chen et al. (2021b) | - | - | 66.00/48.88 | 75.06/59.87 | - | - |
| XLM-R | - | - | 64.14/46.00 | 73.54/57.55 | 78.48 | 70.28 |
| Info-XLM* | 130M | 94 | 65.85/48.23 | 75.79/59.50 | 79.52 | 72.23 |
| XLM-Align* | 130M | 94 | 65.66/48.47 | 75.35/59.10 | 78.87 | 72.66 |
| **Ours** (XLM-R)* | 1M | 4 | 67.34/49.11 | 76.73/60.87 | 80.63 | 73.31 |
| **Ours** (Info-XLM) | 1M | 4 | **69.07/50.81** | **78.22/61.99** | **81.13** | **73.87** |

Table 2: Average evaluation results on four datasets. #data denotes the amount of pre-training data. #languages represents the language diversities in pre-training data. The results of our model are averaged over 5 runs. * denotes the model build upon of XLM-R. The results of each language are represented in the Appendix E.

We represent two main results in *full-data* setting and *few-shot data* setting. That is, our model utilizes all data in training set to train the model in *full-data* setting. In contrast, the model is fine-tuned with smaller training sets in *few-shot data* setting.

**xMRC Results** Table 2 shows the comparison between our approach and representative systems on four datasets. Specifically, our models surpass these baselines by a large margin on two xMRC datasets. For examples, our model initialized from Info-XLM achieve 69.07%/50.81% (vs. 65.85%/48.23%) in terms of F1/EM score. Compared with Info-XLM or XLM-Align, which both build on top of XLM-R with 130 million data across 94 languages, our model initialized from XLM-R obtains better performances like 67.34%/48.91% (vs. 65.85%/48.23%) on MLQA dataset. It is worthwhile to indicate that our model only pre-tained with 1 million parallel data across 4 languages. The results bear out the effectiveness of our proposed methods even if with limited pre-training data across several languages (1M≪130M).

Also when looked into the Table 3, our models dramatically improve performances in the more challenging *few-shot data* setting. Given a small training set with only 512 examples, our model built on top of XLM-R improve the baseline by 24%/20.4% and 20.59%/18% in terms of F1/EM score on both two datasets. The results show that our method presents a new paradigm for xSL tasks in *few-shot data* setting.

**xNER Results** As presented in Table 2, frist compared with baseline XLM-R, our resulting model on top of XLM-R achieves an average F1 score gain of 1.15% and 3.03%. Then compared with Info-XLM and XLM-Align on top of XLM-R which pre-trained on 130 million large corpus across 94 languages, ours still show superior performances. That is, with less than one percent of the corpus they used, our approach has shown its full potential.

## 5 Ablation Study and Analysis

In this section, we first discuss the importance of each key component in our model. Then we conduct an analysis of alleviating pretrain-finetune discrepancy by ours and independent contribution of the proposed CLISM at *few-shot data* seting. We further analyse the relation between pre-trainig data and our methods in the Appendix G.

**Key components** We perform an ablation study to better understand the independent contributions of the proposed pre-training schemes. Table 4 shows the performances when removing each key component of our model on MLQA and XQUAD datasets. Obviously, when removing CLISM, the model performances drop dramatically in both two datasets, showing the effectiveness of the proposed pre-training task. Meanwhile, we observe that with CACR or MLM also giving performance improvement, proving each of them plays an important role in our pre-training scheme.

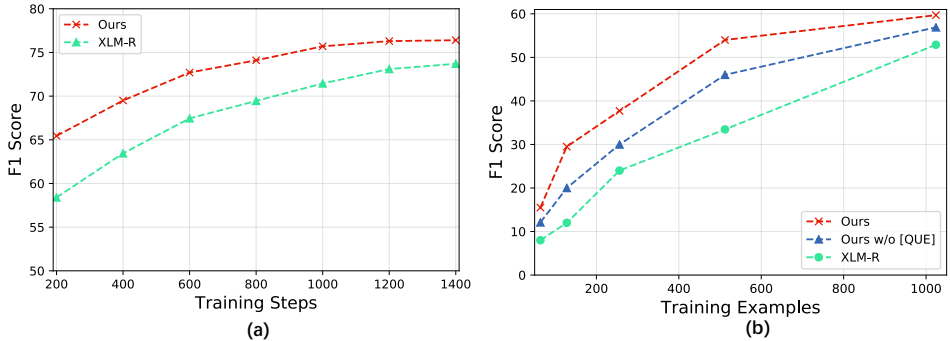

Figure 4: (a): Average F1 score of both Ours and base architecture XLM-R on the XQUAD test set at different training steps. (b): Average F1 score of all methods with giving different few-shot training examples on the XQUAD test set. It is worthwhile to notice that ours denotes the resulting model build on top of XLM-R in these experiments.

| Models | MLQA | XQUAD |
|---|---|---|
| *64 examples* | | |
| XLM-R | 6/1.5 | 8/2 |
| Info-XLM | 7/3.2 | 8.7/5.1 |
| **Ours** (XLM-R) | **14/7.2** | **15.5/8.2** |
| *128 examples* | | |
| XLM-R | 10/3.5 | 12/4 |
| Info-XLM | 11.4/4.5 | 15.7/7 |
| **Ours** (XLM-R) | **26.4/15.9** | **29.5/19.7** |
| *512 examples* | | |
| XLM-R | 26.8/15.5 | 33.35/19.1 |
| Info-XLM | 29.7/18.9 | 36.0/21.0 |
| **Ours** (XLM-R) | **50.8/35.9** | **53.94/37.1** |

Table 3: Model performances across xMRC datasets when the number of training examples is 64, 128 and 512. The more comprehensive results in few-shot data setting are described in Appendix F.

| Algorithms | MLQA | XQUAD |
|---|---|---|
| **Ours** (XLM-R) | **67.34/49.11** | **76.73/60.87** |
| w/o CLISM | 64.56/46.80 | 74.34/58.43 |
| w/o CACR | 66.40/48.01 | 75.87/59.95 |
| w/o MLM | 66.18/47.89 | 75.34/59.61 |

Table 4: Ablation study of pre-training schemes.

**Analysis of alleviating pretrain-finetune discrepancy** We conduct extra experiments to verify the proposed methods whether alleviate the pretrain-finetune discrepancy in cross-lingual sequence labeling tasks. As shown in the Figure 4 (a), in the early stage of training, ours manages to main an absolute lead of 5%-15% on F1 score over XLM-R consistently, showing the resulting model not only speed the fine-tuning time but also achieve superior performances. This demonstrates representations learned by ours are more suitable for data distribution in xSL tasks, leading to better performance.

**Independent Contribution of CLISM** Moreover, in order to understand the independent contribution of the proposed CLISM at few-shot data setting. Figure 4 (b) gives the performance of individual models when giving {64, 128, 256, 512, 1024} examples. It is evident that our model outperforms baseline by large margins. Interestingly, we conduct the following experiments to verify effectiveness of using the [QUE] token during fine-tuning: We don't append [QUE] in the input sequence when fine-tuning our model in down-stream tasks, and thus, we can not utilize its representation as *question*. Thereafter, we follow the standard setting (Liang et al., 2021a; Chen et al., 2021b) to fine-tune ours in xSL tasks. Nevertheless, as shown in the 4 (b), the model performance drops heavily especially at smaller training data setting, such as 64, 128 examples. We interpret this phenomenon as follows: After pre-training, the [QUE] token captures much information about question that can be used to select the span from context in xSL tasks.

## 6 Conclusion

In this paper, heading for bridging the discrepancy including the objective and representation gap between pre-training and fine-tuning stages in cross-lingual sequence labeling tasks. We propose two pre-training strategies: cross-lingual language informative span masking (CLISM) and ContrAstive-Consistency Regularization (CACR), reaching surprisingly superior results on various benchmarks even when only a hundred examples available. We further prove that even with less than a hundredth of the data needed to train other xPLMs, our model still achieves better performances. As an extension of our future work, we will explore how to apply our methods to other nlp tasks.

# Appendix

## A Broader Impacts

The main contributions of this paper are towards tackling fundamental issue during pre-training and fine-tuning in xSL tasks when leveraging pre-trained language models. The proposed method alleviates the discrepancy between pretrain-finetune paradigm, and thus improving the results of xPLMs in these tasks. Therefore, ideas of this paper can be applied in xPLM-based application systems such as cross-lingual question answering systems or robots. More generally, we expect the core idea of this paper can give insights of other research communities who want to build a more robust and effective xPLM-based model in their down-stream tasks. Admittedly, the proposed strategies are restricted with cross-lingual sequence labeling tasks, and biases in training dataset also influence the performance of the resulting model. These concerns warrant further research and consideration when utilizing this work to build cross-lingual application systems.

## B Informative Span Selection

As illustrated in the paper Section 3.1, it is essential to select semantically meaningful spans before pre-training. And some extracted entities could be unreasonable due to the performance limit of off-the-shelf NER tools. Hence, we make several rules to filer out some recurring spans which tend to be uninformative:

- Spans which only consist of stop words will be filtered out.
- The boundary of spans must be the words.
- The max sequence length of each span is limited to 10.

.

## C Fine-tuning Settings

**Fine-tuning Details** We train our model on two xSL tasks: xMRC and xNER. For both two tasks, we utilize Adam optimizer to train the model and batch size is set to 32. For xMRC, the learning rate is 3e-5, the maximum sequence length is set to 384. We train our model using $8 \times$V100-32G GPUs with 5 epochs for fine-tuning. For xNER, the learning rate is set to 5e-5, the max length is 128. It takes 5 epochs with using $8 \times$V100-32G GPUs to get the best checkpoint of our model.

**Few-shot Data Details** We also conduct few-shot data experiments on downstream datasets in xMRC. For each, dataset, we random sample few-shot training datasets from the original full-data sets, with the sampling size $\{64, 128, 256, 512, 1024\}$. To reduce variance, for each few-shot training dataset size, we random sample 5 times with different seeds, and report average results. When training our model on few-shot datasets, we choose 200 steps for optimizing the resulting model.

## D xSL datasets

**Cross-Lingual Machine Reading Comprehension** MLQA and XQUAD are two popular xMRC benchmarks, which provide train dataset in English and test datasets in multiple low-resouce language. In this work, we evaluate our methods on six languages: including *English, Arabic, German, Spanish, Hindi, Vietnamese*.

**Cross-lingual Name Entity Recognition** CoNLL and WikiAnn are common xNER benchmarks. All of those datasets are annotated with 4 types of entities, namely PER, LOC, ORG and MISC in BIO tagging schema following (Pan et al., 2017). Of note, CoNLL and WikiAnn both contain four language test sets: *Spanish, Dutch, English, German*. Besides, WikiAnn additional consists of another two language test sets: *Hindi, Arabic*

## E Detailed Results on cross-lingual sequence labeling tasks

### E.1 Results on MLQA

Table 5 shows results of our models in each language.

### E.2 Results on XQUAD

Table 6 presents results of our models in each language on XQUAD.

### E.3 Results on CoNLL

Table 7 presents results of our models in each language on CoNLL.

### E.4 Results on WikiAnn

Table 8 presents results of our models in each language on CoNLL.

## F Few-shot results on xMRC datasets

We present our model performance with different number of training examples:

| Models | en | es | de | ar | hi | vi | Avg. |
|---|---|---|---|---|---|---|---|
| **Ours**(XLM-R) | 69.14/47.1 | 58.66/37.79 | 65.43/48.80 | 78.67/65.23 | 69.39/49.31 | 62.72/42.07 | 67.34/49.11 |
| **Ours**(Info-XLM) | 69.94/48.61 | 61.66/40.79 | 67.03/50.79 | 79.98/66.73 | 70.90/52.31 | 64.72/45.57 | 69.07/50.81 |

Table 5: The performance of our models on MLQA datasets.

| Models | en | es | de | ar | hi | vi | Avg. |
|---|---|---|---|---|---|---|---|
| **Ours**(XLM-R) | 80.2/66.5 | 76.0/60.01 | 75.7/61.62 | 68.97/50.93 | 68.11/50.67 | 74.12/52.52 | 76.73/60.87 |
| **Ours**(Info-XLM) | 80.98/68.61 | 77.46/61.35 | 74.53/61.19 | 71.61/54.28 | 70.40/52.63 | 73.72/51.69 | 78.22/61.99 |

Table 6: The performance of our models on XQUAD datasets.

$\{64, 128, 256, 512, 1024\}$, shown in Table 9

## G  Pre-training data for CLISM

In this component, we further conduct extensive experiments to explore the relation between pre-training data and model performances. As shown in Figure 5, we present the resulting model performances on xMRC datasets with different number of pre-training data: $\{1M, 2M, 4M, 8M, 12M\}$. Obviously, ours achieve best results given 1 million pre-training data. With the amount of pre-training data increased, the performances drop gradually due to overfitting. The results prove the effectiveness of our proposed pre-training paradigm, that is, our methods could optimize the xPLMs sufficiently even if with limited pre-training data.

| Models | en | es | nl | de | Avg. |
|---|---|---|---|---|---|
| **Ours**(XLM-R) | 91.1 | 78.3 | 78.0 | 70.5 | 80.63 |
| **Ours**(Info-XLM) | 91 | 77.6 | 80 | 71.9 | 81.13 |

Table 7: The performance of our models on CoNLL datasets.

| Models | en | es | de | ar | hi | nl | Avg. |
|---|---|---|---|---|---|---|---|
| **Ours**(XLM-R) | 81.88 | 77.7 | 80 | 74.33 | 57 | 68.95 | 73.31 |
| **Ours**(Info-XLM) | 82.5 | 71.7 | 80.5 | 76.3 | 63 | 69.5 | 73.87 |

Table 8: The performance of our models on WikiAnn datasets.

| Models | MLQA | XQUAD |
|---|---|---|
| *64 examples* | | |
| XLM-R | 6/1.5 | 8/2 |
| Info-XLM | 7/3.2 | 8.7/5.1 |
| **Ours** (XLM-R) | **14/7.2** | **15.5/8.2** |
| *128 examples* | | |
| XLM-R | 10/3.5 | 12/4 |
| Info-XLM | 11.4/4.5 | 15.7/7 |
| **Ours** (XLM-R) | **26.4/15.9** | **29.5/19.7** |
| *256 examples* | | |
| XLM-R | 18.1/9.8 | 20.4/10.4 |
| Info-XLM | 19.8/10.5 | 26.2/13.4 |
| **Ours** (XLM-R) | **39.3/25.8** | **44.2/29.0** |
| *512 examples* | | |
| XLM-R | 26.8/15.5 | 33.35/19.1 |
| Info-XLM | 29.7/18.9 | 36.0/21.0 |
| **Ours** (XLM-R) | **50.8/35.9** | **53.94/37.1** |
| *1024 examples* | | |
| XLM-R | 38.8/23.0 | 43.25/28.03 |
| Info-XLM | 42.7/24.7 | 48.23/30.2 |
| **Ours** (XLM-R) | **57.8/42.4** | **61.25/44.92** |

Table 9: Model performances across xMRC datasets when the number of training examples is $\{64, 128, 256, 512, 1024\}$.

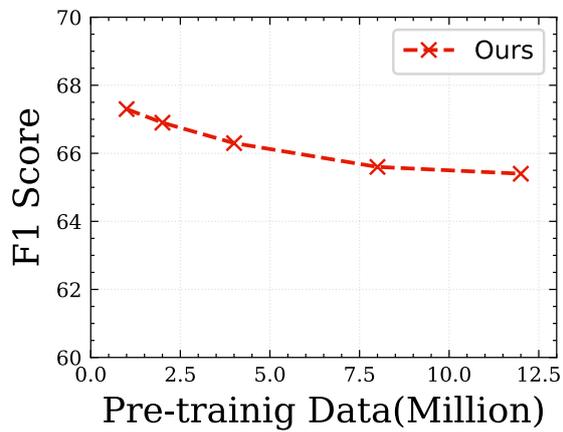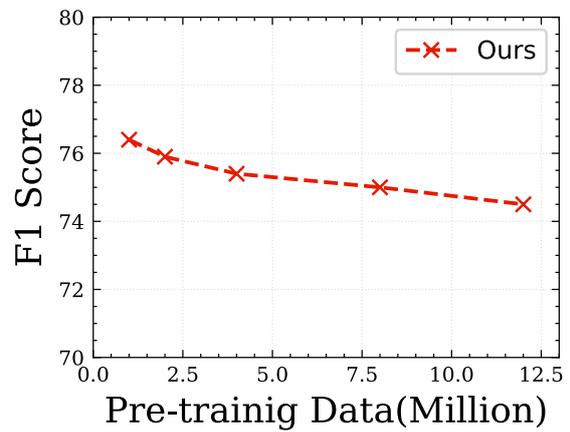

(a) MLQA  (b) XQUAD

Figure 5: Model performance with different pre-training data on MLQA and XQUAD datasets.